# A Geometry-Informed Deep Learning Framework for Ultra-Sparse 3D Tomographic Image Reconstruction

Liyue Shen*, Wei Zhao*, Dante Capaldi, John Pauly, and Lei Xing

**Abstract**— Deep learning affords enormous opportunities to augment the armamentarium of biomedical imaging, albeit its design and implementation have potential flaws [1], [2], [3]. Fundamentally, most deep learning models are driven entirely by data without consideration of any prior knowledge, which dramatically increases the complexity of neural networks and limits the application scope and model generalizability. Here we establish a geometry-informed deep learning framework for ultra-sparse 3D tomographic image reconstruction. We introduce a novel mechanism for integrating geometric priors of the imaging system. We demonstrate that the seamless inclusion of known priors is essential to enhance the performance of 3D volumetric computed tomography imaging with ultra-sparse sampling. The study opens new avenues for data-driven biomedical imaging and promises to provide substantially improved imaging tools for various clinical imaging and image-guided interventions.

**Index Terms**—Geometry-informed deep learning, Sparse-view 3D image reconstruction, Deep learning, Image reconstruction, Tomographic imaging.

——————————— ◆ ———————————

## 1 INTRODUCTION

MODERN medicine and biomedical research rely heavily on the information from various imaging modalities such as X-ray computed tomography (CT), positron emission tomography (PET), magnetic resonance imaging (MRI), ultrasound imaging, and photoacoustic imaging, for diagnosis and medical interventions. In the U.S., for example, around 80 million CT scans, 40 million MRI scans and 2 million PET scans are performed each year, and the imaging data are employed to guide the entire course of the patients' treatment. The demand for better and fast imaging techniques continue to rise over the years in both clinical practice and biomedical research. In general, the task of volumetric imaging is to create a three-dimensional (3D) representation of an object or body from a set of sensor measurement data, which presents an intermediate representation of the object via spatial encoding of imaging content. A common strategy to transform the sensor data acquired at a lower dimension (e.g., in CT and PET) or a different data domain (e.g., in MRI) to 3D images is through inversion of the encoding function. In unfolding the sensor data to place the right image content to the right voxel, traditional analytical or model-based image reconstruction methods, such as iterative reconstruction based on spatial grid, are deployed and used as the backbone during the reconstruction calculation. Despite their enormous success, traditional approaches are susceptible to noise, motion artifacts and missing data, thus fail to yield high-fidelity images in sparse sampling when the Shannon-Nyquist theorem is seriously violated [4].

Recent advances in deep learning [5], [6] enable data-driven image reconstruction by training deep neural networks to fit the function mapping the input sensor mensurements to the target image, which provide unique opportunities to overcome the limitations of traditional approaches. Up to now, although deep learning has demonstrated impressive performance in image reconstruction and recognition [7], [8], [9], [10], [11], [12], the model prediction is entirely driven by the training on large-scale dataset. The performance of learning-based model depends on various factors including training data distribution, network structure as well as hyper-parameters, leading to resultant models with little transparency and interpretability [13]. Especially, for deep learning-based image reconstruction, an underlying deficiency responsible for many of the shortcomings is the lack of explicit geometric guidance in the data-driven model. In the traditional approaches, the image reconstruction calculation is performed on a predefined grid that integrates by default the spatial information in image reconstruction. In most existing data-driven techniques, reconstruction of the spatially distributed image content is learned entirely from training data through features extraction, which may result in geometric misalignment such as in multi-view image processing [14], especially when the training data are imperfect (e.g. having artifacts, noise, or other uncertainties). To accurately reconstruct the 3D images, sophisticated deep neural networks are needed to understand and disentangle the spatial transformation and image content information embedded in the training data [15], which hinders the acceptance of data-driven approaches in many practical applications.

• L. Shen, W. Zhao, D. Capaldi, L. Xing, and J. Pauly are with Stanford University, Stanford, CA, 94305, USA. E-mail: liyues@stanford.edu, zhaow85@stanford.edu, dcapaldi@stanford.edu, pauly@stanford.edu, lei@stanford.edu
• * These authors contributed equally to this work.

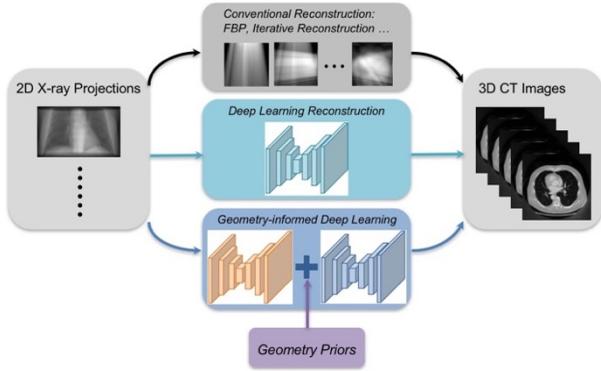

Fig. 1. Different 3D tomographic image reconstruction methods. Among different categories of CT reconstruction algorithms, the geometry-informed deep learning approach not only learns the mapping function driven by large-scale training data, but also integrates geometric priors to bridge the dimensionality gap between the 2D projection domain and 3D image domain.

Here, we introduce an effective strategy of incorporating the prior geometric relationship between the input and output data domains to augment the data-driven learn-based tomographic x-ray imaging. The proposed Geometry-Informed Image Reconstruction (GIIR) relieves the burden for the model to gain comprehension of the system geometry and allows the model to focus on learning other sematic unknowns. We show that the approach makes high-fidelity data-driven image reconstruction possible, even in the limit of ultra-sparse sampling where the most sophisticated iterative reconstruction [16], [17] with regularization fail to yield artifact-free images. We also demonstrate the proposed GIIR model outperforms current data-driven deep learning approaches without geometry priors (Fig. 1).

Generally, in tomographic 3D CT imaging, the 2D projection x-ray measurements (Fig. 1) are the sensor data that encode the internal anatomy of the 3D subject, with the encoding function determined by the physics of the x-ray and media interaction. Figure 2 shows the overall framework of the proposed GIIR model and illustrates the way that the known imaging geometry is integrated into the dual-domain deep learning. Specifically, GIIR framework consists of three modules: a) a 2D projection generation network (2D-Net) (Fig. 3) learns to generate novel-view projections from the given sparse views; b) a geometric back-projection operator (Fig. 4) transforms the 2D projections to 3D images, referred as geometric preserving images (GPIs), which geometrically relates the pixelated 2D input data to the corresponding ray-lines in 3D space; and c) a 3D image refinement network (3D-Net) (Fig. 5) learns to refine the GPIs to reconstruct the final 3D images.

In GIIR model (Fig. 2), the 2D-Net and 3D-Net are trained to learn how to complete the missing information in 2D projection domain and 3D image domain, respectively. The back-projection operator provides the underlying geometry link between the 2D and 3D image domains without any learned parameter. In this way, the information encoded in the sparse projections is partially unfolded back to the 3D image space deterministically, which greatly simplifies the overall learning task and facilitates the information flow from 2D to 3D image domains. This strategy allows the network to maximally exploit the information buried in the training data. Practically, GIIR pushes the CT imaging to ultra-sparse limit, which may provide a viable solution for volumetric imaging with significantly reduced imaging dose and simplified hardware design.

## 2 RELATED WORK

In this section, we introduce the previous works in the relative research fields.

### 2.1 Shape Reconstruction in Natural Images

The problem of 3D object reconstruction in natural images or photographic images has been investigated for many years. With recent advances in deep learning, various deep neural networks have been developed to reconstruct the 3D object shapes from single-view [18], [19], [20], [21] or multi-view [22], [23], [24] input images. Various network architectures and training stratgies have been proposed and achieved much progress in reconstructing 3D object surface in different representations such as voxel grid, mesh representation, and point cloud. However, these reconstruction methods can not be directly applied to 3D tomographic reconstruction for modalities like CT, because of the fundamental differences in their imaging mechanisms. In photography imaging, the lights hit the objects' outer surface and thus take an image of the object shape only. In tomographic imaging, the penetrating waves get through the object to take the image of the internal structure. This work aims at investigating the 3D tomographic reconstruction problem, and proposing an efficient learning-based reconstruction approach incorporating corresponding tomographic imaging physics.

### 2.2 Tomographic Reconstruction in Medical Images

Classical tomographic image reconstruction has to meet the requirement of the Shannon-Nyquist sampling theorem to avoid aliasing artifacts. In cases of undersampling scenarios, such as sparse view or ultra-sparse view, there would be severe artifacts if no measure is taken [25]. To address this issue, image reconstruction algorithms using iterative framework have been investigated extensively [26], [27], [28], [29]. In iterative image reconstruction, prior knowledge (i.e., presumed characteristics) is able to be incorporated into the reconstruction process by using a regularization constraint [30] or the maximum a posteriori approach [31]. The prior knowledge can either be the characteristics (e.g., Poisson statistics properties [17], [32]) in the sampling data or the characteristics (piece-wise constant [33]) in the resultant image. Although iterative image reconstruction has the potential to mitigate the image artifacts, especially the artifacts introduced by the low photon statistics, it is still challenging to address aliasing artifacts. Thus, tomographic reconstruction with ultra-sparse sampling remains an open question. This is the reason why modern CT vendors use low tube current instead of sparse view to reduce radiation dose. Meanwhile, it is also a challenge to incorporate complicated prior knowledge into the iterative framework which may result in a nonconvergent objective function.

With the advances of deep learning in recent years,

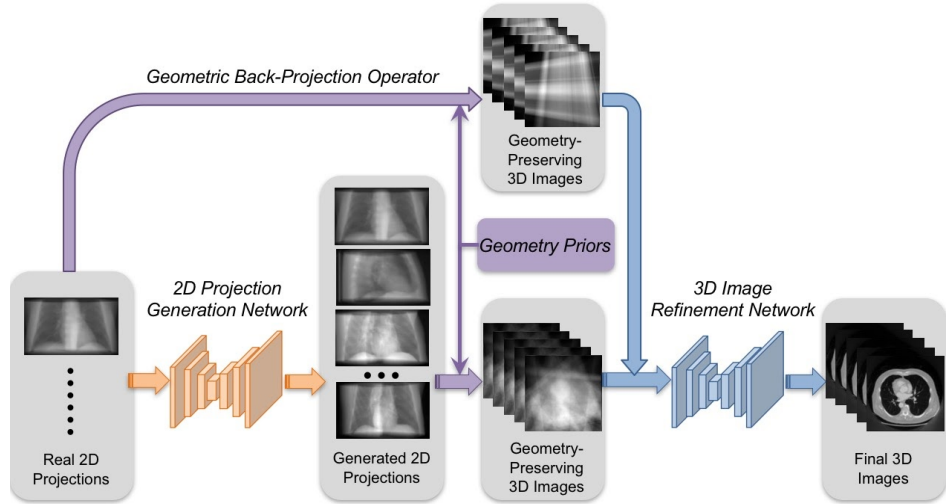

Fig. 2. Geometry-informed image reconstruction (GIIR). Overall framework of GIIR model with dual-domain learning modules of 2D projection generation network (orange arrows), 3D image refinement network (blue arrows) and the geometric back-projection operator (purple arrows) of deterministic geometric transformation.

learning-based tomographic image reconstruction with sparse-view or limited-angle sampling has attracted much attention. Previous studies propose various deep networks for reconstrucing 2D image from limited sinogram by integrating iterative reconstruction with network [34], combining DenseNet and deconvoution to construct a two-step model [35], and using multi-resolution U-Net variant framework [36]. Particularly, [37] [38] introduce dual-domain learning for solving metal artifacts problem in 2D CT reconstruction through sinogram interpolation and image denosing. However, all these previous works focus on the 2D CT reconstruction problem which aims at reconstructing 2D image from 1D projection (sinogram). In this work, we study the 3D CT reconstruction problem to reconstruct 3D volume from 2D projections, where these previous methods are not directly applicable. Besides, [39] maps filtered back-projection to the layer in neural networks for solving the limited-angle reconstruction problem, but it doesn't study for ultra-sparse views how to combine the projection layer with dual-domain learning in both 2D projection and 3D volume domains. Moreover, different from previous works [34], [35], [36] in which the sparse sinogram with 60-120 input views is studied, our work focuses on ultra-sparse-view reconstruction with 1, 2, 3 views. In the ultra-sparse limit, the problem of tomographic image reconstruction is dramatically different.

For low-dose and high-speed imaging applications, a novel strategy of 3D tomographic reconstruction with ultra-sparse sampling data is urgently needed. Given the powerful learning ability of neural networks, deep learning provides an effective way to incorporate various complex prior knowledge to facilitate ultra-sparse image reconstruction. Although most recent works [11], [12] have started using deep learning models to directly learn the mapping from sparse-view projections to tomographic images, these methods do not consider the geometry priors from tomographic imaging system, and thus suffer from the potential generalizability and interpretability problems by only replying on data-driven. Therefore, this work investigates how to incorporate geometry priors into deep learning models to address these questions and proposes a more effetient framework for ultra-sparse-view 3D CT reconstruction.

### 2.3 Geometry-Informed Deep Learning

Combining geometric priors with deep learning models [40] has raised more attention in computer vision and artificial intelligence. Recent study [15] finds that the current state-of-the-art deep learning methods for single-view object shape reconstruction may not actually perform image reconstruction but likely image classification. Thus, it raises the question: based on the pure data-driven approach, how can the deep network perform non-trivial reasoning about = 3D object structure of the output space? [15] There is yet no obvious explanation due to the difficulty in interpreting networks. By leveraging underlying physical knowledge, integrating geometry priors with deep learning may be a promising solution to address this question.

But how to combine geometry and deep learning is still an open question in image reconstruction and other tasks. To achive this, some attempts have been made in recent works [14], [18] in computer vision. For examples, Wu et al [18] showed that the symmetry and shading characteristics of human face can be leveraged to reconstruct face shapes from a single image. Tung et al [14] used the egomotion stabilization to match geometry landmark in different views for 3D object detection. In computer graphics, recent studies on neural rendering [41], [42], [43], [44], [45] also demonstrated a trend to incorporate classic volume rendering geometry with neural networks to render high-resolution images. However, due to the fundamental difference in imaging physics, to develop a geometry-informed deep learning approach for tomographic reconstruction requires further investigations. In this work, we present a novel approach to combine the imaging system geometry and deep learning for tomographic image reconstruction.

## 3 OUR APPROACH

The overall workflow of the proposed GIIR framework is demonstrated in Fig. 2. In the first step, the 2D-Net takes the input projections to generate new projections in other view angles. The input projections and the newly generated projections are then independently back-projected to the 3D space to yield two sets of GPIs, which relates the 2D pixels in the projections to their corresponding ray lines in 3D space. The subsequent 3D-Net takes the two GPIs as input to reconstruct the final 3D volumetric images. In addition to the information from input 2D projections, the GPIs also inform the 3D-Net with the underlying imaging system geometry through unfolding the pixel-wise intensity of the 2D projections back to the corresponding voxels in the 3D image space through ray tracing. Following, we will introduce problem formulation and architecture details.

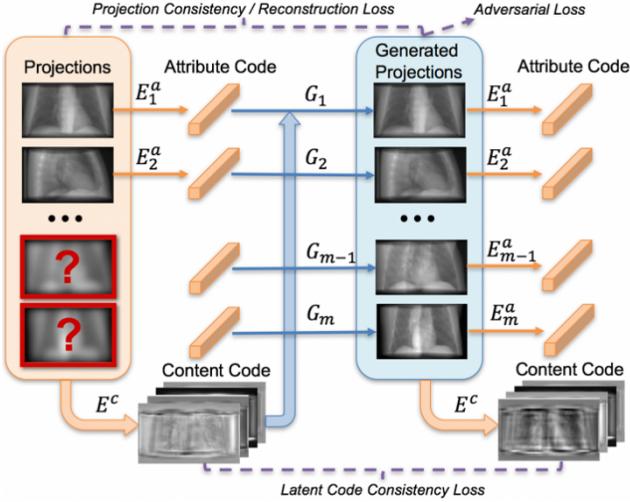

Fig. 3. 2D projection generation network (2D-Net) learns to generate novel views given sparse views (for illustration, two projections are given while unknown views are marked with a red box in inputs). Details of latent codes and loss functions are described in Section 3.2.

### 3.1 Problem Formulation

Firstly, we cast the inverse problem of 3D image reconstruction from 2D projection(s) into a data-driven framework. Given a sequence of 2D projections denoted as $\{p_1, p_2, \cdots, p_n\}$, where $p_i \in \mathbb{R}^{H_{2D} \times W_{2D}}$ ($1 \leq i \leq n$) and $n$ is the number of given 2D projections, the goal is to generate a volumetric 3D image $I$ representing the internal anatomy of the subject. With the 2D projections as input, the deep learning model outputs the predicted 3D volume denoted as $I_{pred} \in \mathbb{R}^{C_{3D} \times H_{3D} \times W_{3D}}$, where $I_{truth} \in \mathbb{R}^{C_{3D} \times H_{3D} \times W_{3D}}$ is the ground truth 3D image serving as the reconstruction target. Note that network prediction $I_{pred}$ is of the same size as ground truth image $I_{truth}$, where each entry is a voxel-wise intensity. Thus, the problem is formulated as finding a mapping function $\Phi$ that transforms 2D projections to 3D images. To solve this problem, we develop a GIIR framework to learn the mapping function $\Phi$. As aforementioned, GIIR consists of three modules: 1) 2D-Net $\phi_1$ with the network weights denoted as $W_1$, 2) geometric back-projection operator $\phi_2$ with the embedding geometry as the known function parameters $W_2$, and 3) 3D-Net $\phi_3$ with the network weights denoted as $W_3$. Thus, the 2D-3D reconstruction process can be formulated as:

$$I_{pred} = \Phi(p_1, p_2, \cdots, p_n)$$

$$= \phi_3(\phi_2(\phi_1(p_1, p_2, \cdots, p_n; W_1); W_2); W_3) \quad (1)$$

In the following, we introduce the implementation details of the three modules $\phi_1$, $\phi_2$, $\phi_3$ in order.

### 3.2 2D Projection Generation Network (2D-Net)

We develop a multi-view X-ray projection synthesis model to generate 2D projections at novel-view angles from sparse input views. Suppose there are $m$ view angles: $\chi_1, \chi_2, \cdots, \chi_m$. Let $p_1 \in \chi_1$, $p_2 \in \chi_2, \cdots, p_m \in \chi_m$ be the projections from $m$ different view angles. $\{p_1, p_2, \cdots, p_m\}$ is a set of paired multi-view projections that depict the underlying imaging subject from different view angles. For each sample, we assume $n$ projections are given as input source views. The goal here is to generate the other $(m - n)$ projections at target view angles and complete missing information in the 2D projection space through deep learning.

To proceed, we assume that the multi-view projections share some latent representations of the underlying imaging subject such as the anatomy structural information, which is named as "content code". Meanwhile, projections at different view angles also contain the view-dependent attributes, named as "attribute code", which is corresponding to the rotation-related characteristics in projections at different view angles. In this way, the novel-view projections could be generated by combining the content code and attribute code. Based on this assumption and motivated by the representational disentanglement in [46], [47], we built a 2D projection generation network for multi-view X-ray projections synthesis. Note that models in [46], [47] were developed for multi-domain image translation problem such as multi-contrast MRI images, which are assumed to have shared content information and domain-specific style information corresponding to different imaging contrasts. In X-ray projection generation, the novel-view projection is synthesized by combining the across-view structural information from input source views and view-dependent attributes from target view angles.

Specifically, as demonstrated in Fig. 3, the 2D-Net mainly consists of three sub-modules: 1) Across-view content encoder $E^c$: encodes anatomic structural information shared by projections from different view angles $E^c(p_1, p_2, \cdots, p_m) = c$ (bold orange arrows); 2) View-dependent attribute encoder $E_i^a$ ($1 \leq i \leq m$): encodes information in the projections which is exclusively corresponding to different view angles $E_i^a(p_i) = a_i$ (orange arrows); and 3) Projection generator $G_i$ ($1 \leq i \leq m$): generates novel-view projections by combining across-view and view-dependent information learned from encoders $G_i(c, a_i) = \hat{p}_i$ (blue arrows). The proposed model learns to extract and disentangle the across-view anatomy information and view-dependent rotation attributes from projections at different view angles through a data-driven deep learning approach. Similar to [46], [47], we assume the prior distribution for attribute latent code is standard Gaussian distribution $\mathcal{N}(0, I)$ to capture the distribution of rotation characteristics. During training, the attribute codes of target views are sampled from the prior distribution. For inference, attribute code is fixed and combined with anatomic content to generate target-view projections.

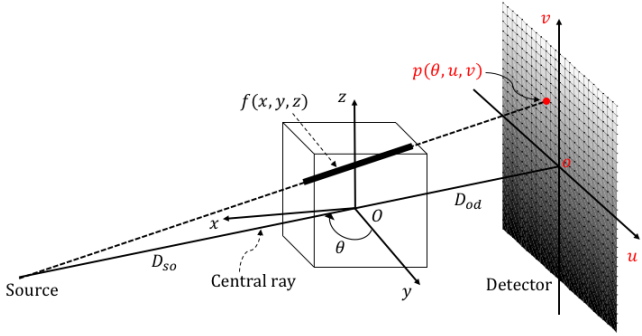

Fig. 4. Geometric back-projection operator unfolds 2D projections to 3D space geometrically to generate geometry-preserving 3D images.

Specifcially, the training objectives contain cycle-consistency loss, adversarial loss and reconstruction loss on the generated projections. Firstly, for input source-view projections, the generated projections after the encoding and decoding should recover the original projections. Thus, projection consistency loss adds such a constrain in the cycle of "Projection → Code → Projection" (Fig. 3).

$$\mathcal{L}_{cyc}^{p_i} = \mathbb{E}_{p_i \sim \mathcal{P}(p_i)} \left[ \| G_i\big(E^c(p_1,\cdots,p_m), E_i^a(p_i)\big) - p_i \|_1 \right] \quad (2)$$

where $\mathcal{P}(p_i)$ is the projection distribution. Likely, the latent codes should also keep consistent in the cycle of "Code → Projection → Code" (Fig. 3). Thus, the latent code consistency loss can be formulated as follows:

$$\mathcal{L}_{cyc}^{c} = \mathbb{E}_{c \sim \mathcal{P}(c),\, a_i \sim \mathcal{P}(a_i)} \left[ \| E^c\big(G_1(c, a_1),\cdots,G_m(c, a_m)\big) - c \|_1 \right]$$
$$\mathcal{L}_{cyc}^{a_i} = \mathbb{E}_{c \sim \mathcal{P}(c),\, a_i \sim \mathcal{P}(a_i)} \left[ \| E_i^a\big(G_i(c, a_i)\big) - a_i \|_1 \right] \quad (3)$$

where $\mathcal{P}(a_i)$ is the assumed prior distribution of attribute code, which captures various view-dependent characteristics related to different view angles. The content code is sampled from $\mathcal{P}(c)$ by firstly sampling projection distributions $p_i \sim \mathcal{P}(p_i)$ ($1 \leq i \leq m$) and then getting through content encoder $c = E^c(p_1, p_2, \cdots, p_m)$. To be specific, distribution $\mathcal{P}(c)$ describes various anatomy structures across different patients.

In order to enforce the generated target-view projections to resemble the ground truth projections, we add the reconstruction loss for different views as follows in the training objective.

$$\mathcal{L}_{rec}^{p_i} = \mathbb{E}_{c \sim \mathcal{P}(c),\, a_i \sim \mathcal{P}(a_i)} [\| G_i(c, a_i) - p_i \|_1] \quad (4)$$

Moreoever, in recent researches on image generation and reconstruction, adversarial training has shown advantages in providing improved image quality. Therefore, in training the projection synthesis model, we use both reconstruction loss and adversarial loss to improve the image quality of the generated projections. The adversarial loss is defined as follows, with the discriminator $D_i$ to classify between the generated projections and real projections.

$$\mathcal{L}_{adv}^{p_i} = \mathbb{E}_{c \sim \mathcal{P}(c),\, a_i \sim \mathcal{P}(a_i)} \left[ \log\big(1 - D_i(G_i(c, a_i))\big) \right] + \mathbb{E}_{p_i \sim \mathcal{P}(p_i)} \left[ \log\big(D_i(p_i)\big) \right] \quad (5)$$

To sum up, the total training objective is shown as follows, with $n$ projections are given as input source views to generate $(m - n)$ target-view projections.

$$\min_{E^c, E_i^a, G_i} \max_{D_i} \mathcal{L}(E^c, E_1^a, \cdots, E_m^a, G_1, \cdots, G_m, D_1, \cdots, D_m)$$
$$= \lambda_{cyc} \left( \sum_{i=1}^{n} \mathcal{L}_{cyc}^{p_i} + \sum_{i=1}^{m} \mathcal{L}_{cyc}^{a_i} + \mathcal{L}_{cyc}^{c} \right)$$
$$+ \sum_{i=1}^{m} \big( \lambda_{rec} \mathcal{L}_{rec}^{p_i} + \lambda_{adv} \mathcal{L}_{adv}^{p_i} \big) \quad (6)$$

where $\lambda_{cyc}$, $\lambda_{rec}$, $\lambda_{adv}$ are hyper-parameters of the loss weights. In experiments, we set $\lambda_{cyc} = 1$, $\lambda_{rec} = 20$, $\lambda_{adv} = 1$.

### 3.3 Geometric Back-Projection Operator

The geometric back-projection is the inverse transformation of the forward-projection in X-ray imaging. As shown in Fig. 4, during X-ray imaging data acquisition, penetrating waves from the source pass through the imaging subject and project onto the detector plane by integrating the intersected voxel intensities in the 3D volume. In this way, multiple X-rays provide projection images of the internal anatomic structure of the imaging subject. In the forward-projection, the geometric relationship between the ray lines of incident X-ray beams and the subject content is determined by the geometry of X-ray imaging system, including the distance between source and volume center, the distance between volume center and detector plane, the physical size of subject voxels, the physical size of detector pixels, the cone-beam geometry and the projection view angles. All these information are the relevant geometry priors provided from the X-ray imaging system.

For image reconstruction, we aim at solving the inverse problem to reconstruct the 3D subject volume from 2D projections at different view angles. To relate the 2D projection domain and 3D image domain, we conduct back-projection operation to convert the 2D projections back to the 3D space according to the imaging system geometry as aforementioned. The back-projection operation depends on the same set of geometric relationship as the forward-projection. In this process, the intensity of a pixel on the 2D projection is placed back to the corresponding voxels of the 3D imaging subject located along the path of the ray line that links the pixel and X-ray source. In this way, the relationship between the pixel-wise intensities on 2D projections and voxel intensities in 3D volumes are incorporated into the 3D image reconstruction.

To be specific, we integrate the imaging geometry into deep learning framework by using the back-projection operation $\mathcal{B}$ to put the pixel intensity back to the corresponding projection line through the point of voxel grid to be reconstructed (Fig. 4). In this way, the GPI is constructed simply by aggregating all the projection lines from different view angles. Suppose that the GPI volume to be reconstructed is denoted as $I_{GPI}(x, y, z)$ with $x, y, z$ representing a point in 3D image space based on the 3D coordinate system. Mathematically, the back-projection operation $\mathcal{B}$ can be formulated by the following equation:

$$I_{GPI}(x, y, z) = \mathcal{B}\{\boldsymbol{p}\} = \sum_{\theta \in \Omega} p\big(\theta, u(x, y, \theta), v(x, y, z, \theta)\big) \quad (7)$$

where $\boldsymbol{p}$ is the assemble of input projections $\{p_1, p_2, \cdots, p_n\}$, $\theta$ is the view angle of a specific projection, $\Omega$ is the assemble of all view angles of input projections, $u$ and $v$ are the ray-projected positions in the detector coordinate system, which can be calculated as follows:

$$u(x,y,\theta) = \frac{x\cos\theta + y\sin\theta}{D_{so} + x\sin\theta - y\cos\theta} D_{sd} \quad (8)$$

$$v(x,y,z,\theta) = \frac{D_{sd}}{D_{so} + x\sin\theta - y\cos\theta} z \quad (9)$$

Here $D_{so}$ is the source-to-isocenter distance and $D_{sd}$ is the source-to-detector distance. In this study, we use the geometry of Varian TrueBeam onboard imager with $D_{so}$= 1000 mm, and $D_{sd}$ = 1500 mm. The 3D back-projection operation was implemented using GPU-based parallel computing with CUDA C programming language (see Supplmentary). Using back-projection operation, both the input sparse-view projections and the generated novel-view projections are projected back to 3D image space based on the ray traces, in which the geometry of tomography imaging is integrated to bridge the 2D-3D image domains. Note that the geometric back-projection is a deterministic transformation, which does not need to learn any parameter.

### 3.4 3D Image Refinement Network (3D-Net)

By using the geometric back-projection operator, the two GPIs are produced from the input source-view projections $I_{GPI}^{src}$ and the newly generated novel-view projections $I_{GPI}^{gen}$:

$$I_{GPI}^{src} = \phi_2(p_1, p_2, \cdots, p_n; W_2) \quad (10)$$
$$I_{GPI}^{gen} = (\phi_2(\phi_1(p_1, p_2, \cdots, p_n; W_1); W_2) \quad (11)$$

With two GPIs as inputs, the 3D-Net is trained to complete information in image domain and generate final 3D images:

$$I_{pred} = \phi_3(I_{GPI}^{src}, I_{GPI}^{gen}; W_3) \quad (12)$$

The network architecture is built up on top of a backbone 3D U-Net [48], [49] structure with an encoder-decoder framework and skip connections. In our model, the features of two input GPIs are extracted by using two separate encoder branches. For this purpose, a variant Y-shape model is constructed to learn the image refinement as shown in Fig. 5. Specifically, the Y-shape network is constructed to incorporate the information from both the given source-view projections and the generated novel-view projections. The given input projections are very sparse but all the pixel-wise intensities in the source-view projections are accurate and reliable. Therefore, these data are able to regularize the generated novel-view projections, which may have uncertainty in intensity distribution due to their synthetic nature. It is worthwhile to note that these generated projections could provide a geometry-preserving image with more outstanding anatomic structures of the subject, which simplifies the subsequent learning of image refinement network. Based on these considerations, a Y-shape network structure is constructed for 3D image refinement to reconstruct the final 3D image.

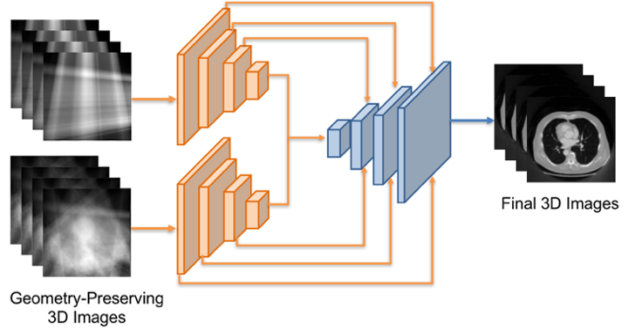

Fig. 5. 3D image refinement network (3D-Net) learns to refine the geometry-preserving 3D images to reconstruct the final 3D image.

To be specific, the encoder consists of four-stage down-sampling blocks to learn hierarchical representations from the input GPIs. Each down-sampling block contains double 3D convolution layers followed by rectified linear unit activation (ReLU) layers [50] and 3D group normalization layers [51]. The down-sampling is implemented by max pooling layer with a step of (2, 2, 2). Asymmetrically, the decoder also consists of four-stage up-sampling blocks to generate final 3D images from the representations learned from the encoder. Each up-sampling block contains double 3D convolution layers with ReLU layers and group normalization layers. We use interpolation function to conduct up-sampling operations. Especially, in order to establish the hierarchical skip connections between two encoders and the decoder, we concatenate the feature maps from both encoders and connect to the corresponding feature level in the decoder, as shown in Fig. 5. In this way, we force the model to utilize information from both the input sparse-view projections and the generated novel-view projections to reconstruct final 3D images. Finally, we use another 3D convolution layer with the kernel size of $1 \times 1 \times 1$, and tangent activation to output the final 3D image with the expected size and data range. The loss function to optimize the 3D image refinement network is:

$$\min_{W_3} \mathcal{L}(W_3) = \mathbb{E}_{I_{GPI}^{src} \sim \mathcal{P}(I_{GPI}^{src}),\ I_{GPI}^{gen} \sim \mathcal{P}(I_{GPI}^{gen})} [\|\phi_3(I_{GPI}^{src}, I_{GPI}^{gen}; W_3) - I_{truth}\|_1] \quad (13)$$

## 4 EXPERIMENTS

In this section, we first introduce the dataset and experimental setting to evaluate the proposed approach for tomography image reconstruction. Then we introduce the comparison methods and training and testing details.

### 4.1 Dataset

To evaluate the effectiveness of the proposed approach, we conduct experiments on a public dataset: The Lung Image Database Consortium and Image Database Resource Initiative (LIDC-IDRI) [52], [53], [54]. The dataset contains 1018 patient cases, each including a volumetric image from a clinical thoracic CT scan. Here, we regard each case as an independent data sample.

In data pre-processing, we resample all the CT images to the same 1 $mm$ resolution in $z$-axis direction and resize cross-sectional images on the $xy$-plane to the size of $128 \times 128$. In experiments, 80% of the data (815 samples) are used for training and validation, while 20% of the data (203 samples) are held out for testing. For each 3D CT image, the 2D projections or digitally reconstructed radiographs (DRRs) in different view angles are obtained by projecting the 3D CT image along respective directions, with the geometry defined by a clinical on-board cone-beam CT imager of TrueBeam system (Varian Medical System, Palo Alto, CA) for radiation therapy [55]. During training, following the standard protocol of data pre-processing, we conduct scaling normalization for both 2D projections and 3D images, where pixel-wise or voxel-wise intensities are normalized to the data range $[-1, 1]$.

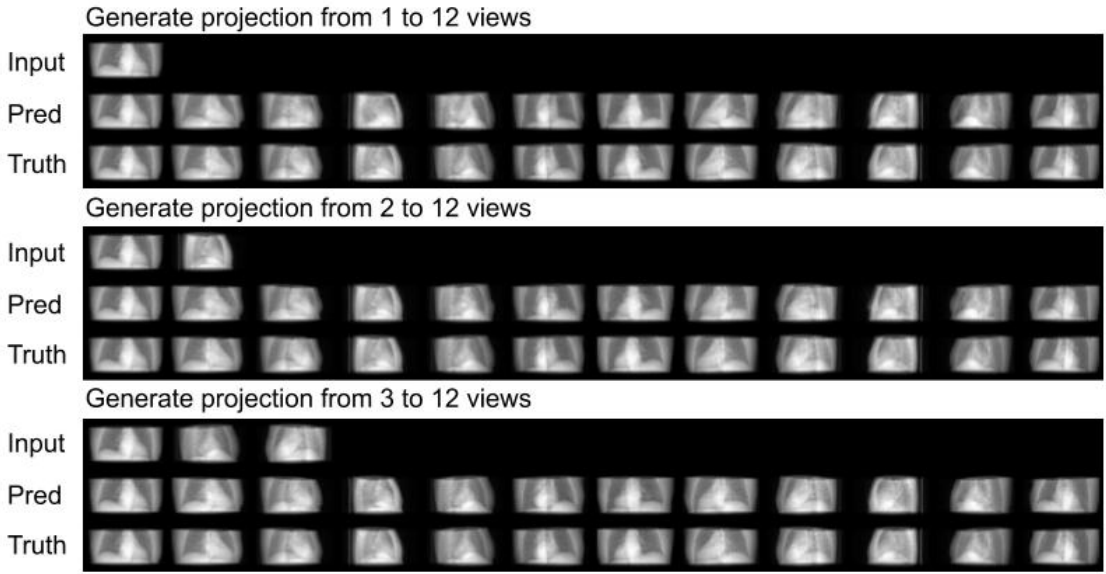

Fig. 6. Results of 2D-Net for generating novel-view projections. Predicted and ground-truth 2D projections (12 angular views evenly distributed over 360 degrees) of a testing sample (with 1-3 input projections respectively). For each group of results, the input projections fed into the network, the predicted novel-view projections output from the network, and the ground-truth of novel-view projections are shown row by row.

### 4.2 Compared Methods

For method comparison, we build a baseline model based on the methodology introduced in [11], [12]. Specifically, the baseline model is a deep learning-based image reconstruction model consisting of a 2D encoder and a 3D decoder. Similar to [11], [12], a feature transformation module in the bottleneck layers connects the 2D and 3D feature space. Note that baseline model is an end-to-end data-driven model without using any geometric prior. For a fair comparison, the baseline model and proposed GIIR model use the same network backbone layers, and are trained with the same dataset and pre-processing procedures.

We conduct experiments of baseline model with 1 to 120 input projections to investigate its learning capability with different number of projections. Multiple input projections are stacked in channel dimension as the input of network. When the number of input projections is more than 30, in order to maximize the network capacity to extract and fuse the information from multiple views, we use a multi-branch 2D encoder to deal with the input projections in multiple groups. For example, when there are 60 input projections, we split them as 2 groups (30 projections in each group) and feed into a two-branch encoder separately. Then the learned representations from multiple encoder branches are concatenated in the bottleneck layer and fed into the 3D decoder. Similarly, a three-branch encoder is used for 90 projections and a four-branch encoder is used for 120 projections. In this way, diverse representations may be learned from different branches and simultaneously contribute to the final image reconstruction. The results of this model will be shown in the next section.

Beyond baseline model, we also compare with the X2CT model introduced in [12], which is especially designed for single-view and two-view reconstruction. Essentially, the model structure of X2CT still follows the encoder-decoder framework without explicitly leveraging geometric priors. The key difference is that X2CT uses a more complicated network architecture with DenseNet blocks and introduces various connections between encoder and decoder as well as between two input views, which is argued as a more efficient way to learn the representation tranfer in feature space [12]. Note that the different network architectures or loss functions can also be adopted in the proposed GIIR framework. Since integration of geometric priors is orthogonal to the network structure development of 2D-Net and 3D-Net, GIIR provides a general model-agnostic approach.

Since the authors of X2CT do not release their code, we reimplement the X2CT model according to the description in [12]. For a fair comparison with baseline and GIIR model, we use the same training dataset, image preprocessing and evaluation metrics in X2CT experiments. Besides, as an ablative study, we conduct experiments to train X2CT model (1) with only reconstruction loss, (2) or with both reconstruction and projection losses. Note that the baseline model and GIIR model are trained with only reconstruction loss.

### 4.3 Training

With the given 2D projections $\{p_1, p_2, \cdots, p_n\}$, the proposed GIIR model aims at predicting the 3D image $I_{pred}$ as close as possible to the ground-truth 3D image $I_{truth}$. The loss functions for 2D-Net and 3D-Net are presented in Sec.3.

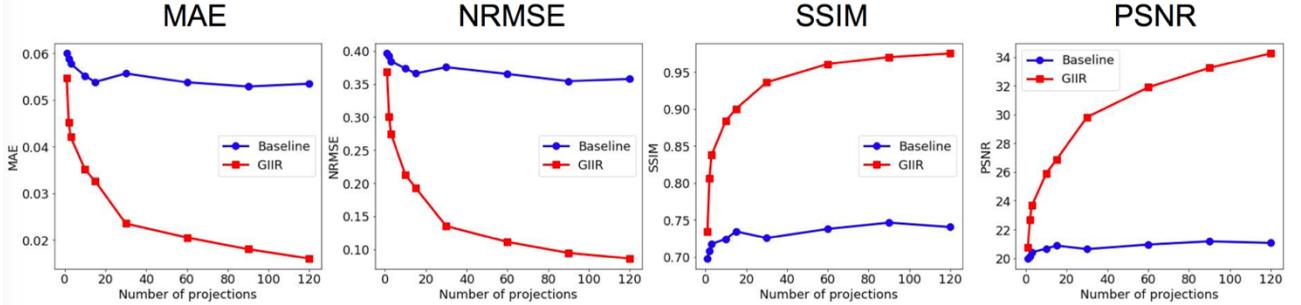

Fig. 7. Evaluation results of "Baseline" model (without inclusion of geometry) sand "GIIR" model (with geometry) for different numbers of input projections. The x axis indicates the number of projections given as inputs. The y axis shows the values of evaluation metrics, MAE, NRMSE, SSIM, and PSNR, respectively. For reference, the MAE, NRMSE, SSIM for the fully sampled ground truth images are 0, 0, 1 respectively, while PSNR is not defined.

The two deep networks are implemented using PyTorch [56] and trained separately by using Adam optimizer[57]. In the training of 2D-Net, we use mini-batch size of 1 and initial learning rate of 0.0001, which is decayed by 0.5 for every 100000 iterations with a total of 110000 iterations. For the training of 3D-Net, mini-batch size is 3 and initial learning rate is 0.0002, which is decayed by 0.2 for every 20000 iterations with a total of 30000 iterations. The baseline model adopts the same training strategy as the 3D-Net. The validation set, which is randomly selected from the training set, is used to tune the hyper-parameters. After finalizing the model structure and hyper-parameters, we use all the training data (815 samples) to train the model and evaluate on the held-out testing set (203 samples). We trained the network using one Nvidia Tesla V100 GPU.

### 4.4 Testing

We evaluate the reconstructed image quality by using both qualitative and quantitative evaluation metrics. Various quantitative metrics are employed to measure the reconstructed 3D images $I_{pred}$: normalized root mean-squared error (NRMSE), structural similarity (SSIM), and peak signal-to-noise-ratio (PSNR). We compute the average values over all testing samples for different metrics respectively. NRMSE measures the difference between prediction and ground-truth images. SSIM score is calculated with an image window and used to measure the overall structural similarity between two images [58]. PSNR gives the ratio in dB between the maximum signal power and noise power in the image. In general, a lower value of NRMSE or a higher SSIM score indicates a better reconstructed image that is closer to the ground-truth. Higher PSNR is always desirable as it implies a better image quality.

In addition to the quantitative metrics as reference, we also display the reconstructed images of testing samples, which present a direct comparison with ground truth and other methods. The same metrics are also used to evaluate the intermediate results of 2D-Net and 3D-Net.

## 5 RESULTS

To evaluate the proposed approach, we conduct experiments on a dataset of 1018 lung CT images of different patients [52]. The 2D-Net is trained by the projection data from different view angles, while the 3D-Net is trained with the paired data of GPIs and ground-truth 3D images.

In our study, the anterior-posterior (AP) projection (0-degree) is used as input for single-view reconstruction. The AP and lateral (90-degree) projections are for two-view reconstruction. When more projections are sampled ($n \geq 3$), they are distributed evenly over the 360 degrees. The target 3D images are the fully sampled CT images reconstructed from 720 projections as used in standard clinical protocol. Following, we demonstrate the results of ultra-sparse view CT reconstruction, as well as intermediate results.

### 5.1 2D Projection Generation

We first evaluate the 2D-Net performance in generating novel-view projections. Experimentally, the network (Fig. 3) is trained to generate 12 projections from 1~3 input projection(s). The choice of the number of generated projections is set in terms of the trade-off between obtaining informative GPIs capturing sufficient object information and avoiding too much error accumulation from projection synthesis. We deploy the trained network on held-out testing dataset. Fig. 6 shows generated projections of a testing sample while results of more testing samples are shown in Supplementary Fig. S1. Quantitative results are reported in Supplementary Table SI.

It is remarkable that 2D-Net could generate novel-view projections that resemble the targets with high fidelity, even given only ultra-sparse input projections. The model generalizes well across different subjects with various anatomy structures as the testing samples shown in Fig. 3 and Supplementary Fig. S1. Results in Supplementary Table SI show that increasing input views can provide more structural information, resulting in more accurate synthesized projections in novel-view angles.

### 5.2 Simplified GIIR for 3D Reconstruction

In order to comprehend the significant role of geometric priors and the functionality of 3D-Net, we first construct a simplified GIIR reconstruction model containing only back-projection operation and 3D-Net without 2D-Net. Specifically, we apply geometric back-projection operator (Fig. 4) to the input 2D projections to produce the GPIs, which then get through the 3D-Net for reconstructing 3D images. Note that, here, all the inputs are ground-truth 2D projections, and the 3D-Net (top half branch of Fig. 2) are trained by using the corresponding GPI from only sthe ground-truth 2D projections.

We conduct experiments using different number of projections (10~120) as input views. With the increasing input projections, the whole model is trained separately

TABLE I
RESULTS OF ULTRA-SPARSE-VIEW RECONTRUCTION

| Reconstruction Methods | Single-view | | | Two-view | | | Three-view | | |
|---|---|---|---|---|---|---|---|---|---|
| | NRMSE | SSIM | PSNR | NRMSE | SSIM | PSNR | NRMSE | SSIM | PSNR |
| Baseline | 0.3961 | 0.6978 | 19.9819 | 0.3927 | 0.7078 | 20.1455 | 0.3841 | 0.7175 | 20.4150 |
| X2CT | 0.3977 | 0.7002 | 20.2308 | 0.3751 | 0.7311 | 20.9120 | - | - | - |
| X2CT(+proj loss) | 0.3766 | 0.7117 | 20.5705 | 0.3596 | 0.7286 | 20.9924 | - | - | - |
| GIIR | 0.3684 | 0.7341 | 20.7699 | 0.3000 | 0.8067 | 22.6868 | 0.2740 | 0.8378 | 23.6688 |

*NRMSE, normalized root mean squared error; SSIM, structural similarity; PSNR, peak signal noise ratio.*

regarding to the specific input projection pattern. During testing, the trained model is deployed under the same setting as training. We plot the averaged evaluation metrics (MAE, NRMSE, SSIM, PSNR) against the number of input projections as the red curve in Fig. 7 denoted as "GIIR" when input projections are more than 10. The resultant images are demonstrated in Supplementary Fig. S2. These results show that, as compared with baseline model results obtained without geometric priors (see next section for details), incorporation of prior geometric information leads to better reconstructed 3D images. This trend continues as the number of projections increases, but the improvement saturates as more and more projections are provided. The "knee" point in the curve appears around 30 projections.

It suggests that, with simplified GIIR, the incorporation of geometric priors could greatly compensate the missing information caused by sparse sampling and yield CT images close to the fully sampled CT (reconstructed from 720 projections), especially when the angular sampling is over 30 projections. But when it comes to more sparse sampling, we observe that the image quality deteriorates gradually as projection number decreases, from Fig. 7 (red curve) and Supplementary Fig. S2. Thus, ultra-sparse view reconstruction needs to be tackled more carefully.

### 5.3 GIIR for Ultra-Sparse 3D Reconstruction
Based on above studies, in order to reconstruct volumetric images with ultra-sparse sampling, either more sophisticated network architecture or additional prior knowledge must be in place. A dual-domain learning framework with incorporation of known geometry priors (Fig. 2) is designed to improve the situation and provide a viable solution for image reconstruction with ultra-sparse sampling. In the proposed GIIR model, two GPIs obtained by back-projecting the ground-truth projections and the generated projections from 2D-Net, are used as inputs to train the two-branch Y-shape 3D-Net simultaneously.

The final results of single-/two-/three-view reconstruction are demonstrated in Fig. 8. The corresponding quantitative evaluations averaged across all testing samples are reported in Table I denoted as "GIIR", which are also plotted in red curve in Fig. 7 when input projections are 1, 2, 3. More testing images are demonstrated in Supplementary Figs. S3-S5. It is worth to note that even with ultra-sparse projections, GIIR model is able to reconstruct 3D CT images with important anatomic structures such as the shape, size and boundary of the human body and organs, as shown in Fig. 8 and Supplementary Figs. S3-S5.

Compared with baseline and X2CT models, the resultant images from GIIR model provide more reliable structures. To be specific, in the testing sample 1 of Fig. S3(a) and Fig. S4(a), GIIR reconstructs more clear shape and shaper boundary for liver and cardio organs, which are closer to ground truth compared with baseline and X2CT results. Similar observation can also be seen in the testing sample 2 shown in Fig. S3(b) and Fig. S4(b). Finally, after averaging across all the testing samples, evaluation metrics reported in Table I give a quantitative comparison between different methods. For a fair comparison, we experiment X2CT model with (1) only reconstruction loss (denoted as "X2CT"), and with (2) both reconstruction and projection losses as introduced in [12] (denoted as "X2CT(+proj loss)"). Note that baseline and GIIR model are trained with only reconstruction loss. To sum up, based on the quantitative and qualitative results comparison, the proposed GIIR method outperforms the baseline and X2CT models in ultra-sparse tomographic image reconstruction.

### 5.4 Method Comparison and Ablative Study
As aforementioned, baseline model is a deep learning-based reconstruction model consisting of a 2D encoder for feature extraction and a 3D decoder for image generation with dimensionality transformation in feature space [11], [12]. As an ablative study to investigate the effectiveness of integrating geometry priors, we develop the baseline model structure by removing the geometric back-projection operation while keeping the 2D encoder in the 2D-Net and 3D generator in the 3D-Net, which use the same backbone layers and training strategy as GIIR model. From the comparison between baseline and GIIR model, we can clearly see the benefit and difference as results of the introduction of geometry priors.

Quantitative evaluations of the baseline model with regard to different numbers of input projections are plotted as the blue curve in Fig. 7 (blue curve) and reported in Table I. From the trend of the curve, it is seen that, when input projections are less than 30, the benefit of adding more views is more pronounced in improving 3D image reconstruction. Beyond 30 projections, only small improvement is observed from providing more input information. Intuitively, a model with more input views should perform better. However, how does a data-driven deep network reasons and realizes this common wisdom, especially in the limit of ultra-sparse views, is not clear. In order to map the input multi-view 2D projections to the output 3D image, the deep network needs to apprehend the cross-dimensionality transformation to transfer the knowledge from 2D projection domain to 3D image domain. In reality, learning such a cross-dimensionality transformation is a highly nontrivial task, especially for a purely data-driven approach. It is conceivable that the challenge is multifaceted. For example, it is challenging for the data-driven algorithm to fuse the extracted semantic features from different input views together without any prior guidance. Considering the memory limit, it seems to be impossible to use a separate 2D encoder for each input projection. Thus, in setting up the baseline model, we design a multi-branch model to adaptively fit different numbers of input views as described in Section 4.2. Even with such a design, as indicated by the results in Fig. 7, direct learning the multi-view feature fusion and 2D-3D feature transformation with a large number of input projections through a data-driven approach is not an easy task. Estblishing a more robust data-driven model via the integration of geometric priors is one of our main motivations of this research. The proposed strategy allows us to combine the knowledge from multiple views with reduced ambiguity and enhance the ultimate success of 3D image reconstruction.

To sum up, the GIIR model with geometry priors outperforms the baseline model without geometry priors. With more than one input projections, the improvement of GIIR model becomes increasingly pronounced. A comparison of the baseline and GIIR models suggests that inclusion of geometric priors greatly facilitates the extraction and fusion of the information from multi-view projections, leading to an improved image reconstruction.

### 5.5 Heatmap Visualization

In order to better understand how geometric priors facilitate the deep learning in 3D image reconstruction, we visualize the intermediate features via heatmap plots. The feature maps are extracted from bottleneck layers of the baseline model and 3D-Net in GIIR. For visualization, multi-channel 3D feature maps are averaged across channel dimension and upsampled to the size of target 3D images.

In Fig. 9, we show cross-sectional slices of 3D feature maps for three testing samples in single-view reconstruction experiments. By integrating geometric transformations, GIIR model gains a comprehension of the systematic structure and spatial relationship. As a result, the deep network can focus on learning the semantic unknowns to improve the image quality and emphasize more on the fine details of the anatomical structures such as the lung, cardio and liver regions. For instance, as shown in testing sample 1 of Fig. 9b, the model highlights the fine structures in lung region such as vascular trees. The highlighted regions of interest are changed adaptively according to the actual lung region across the cross-sectional slices at different locations (pointed by arrows). In testing sample 2 and 3, the model emphasizes the subregions of liver and cardio in consecutive slices (pointed by arrows). Especially, the regions of interest adapt to the actual shape and size of the organ keeping consistent across different slices. However, as shown in Fig. 9a, we cannot see such patterns in the results of baseline model. For the same testing samples, the highlights of heatmaps are more randomly distributed in each slice and inconsistent across different cross-sectional slices. This is not surprising because it is non-trivial for the deep network to reason about the 2D-3D spatial transformation when a purely data-driven approach is used. On the other hand, it is seen that the heatmaps of the GIIR model respect the spatial structure of the underlying subject and keep consistent across sequential cross-sectional slices. To a large extent, in GIIR, the geometric priors relieve the demand for across-dimensionality reasoning in the deep networks, and thus, result in better 3D images.

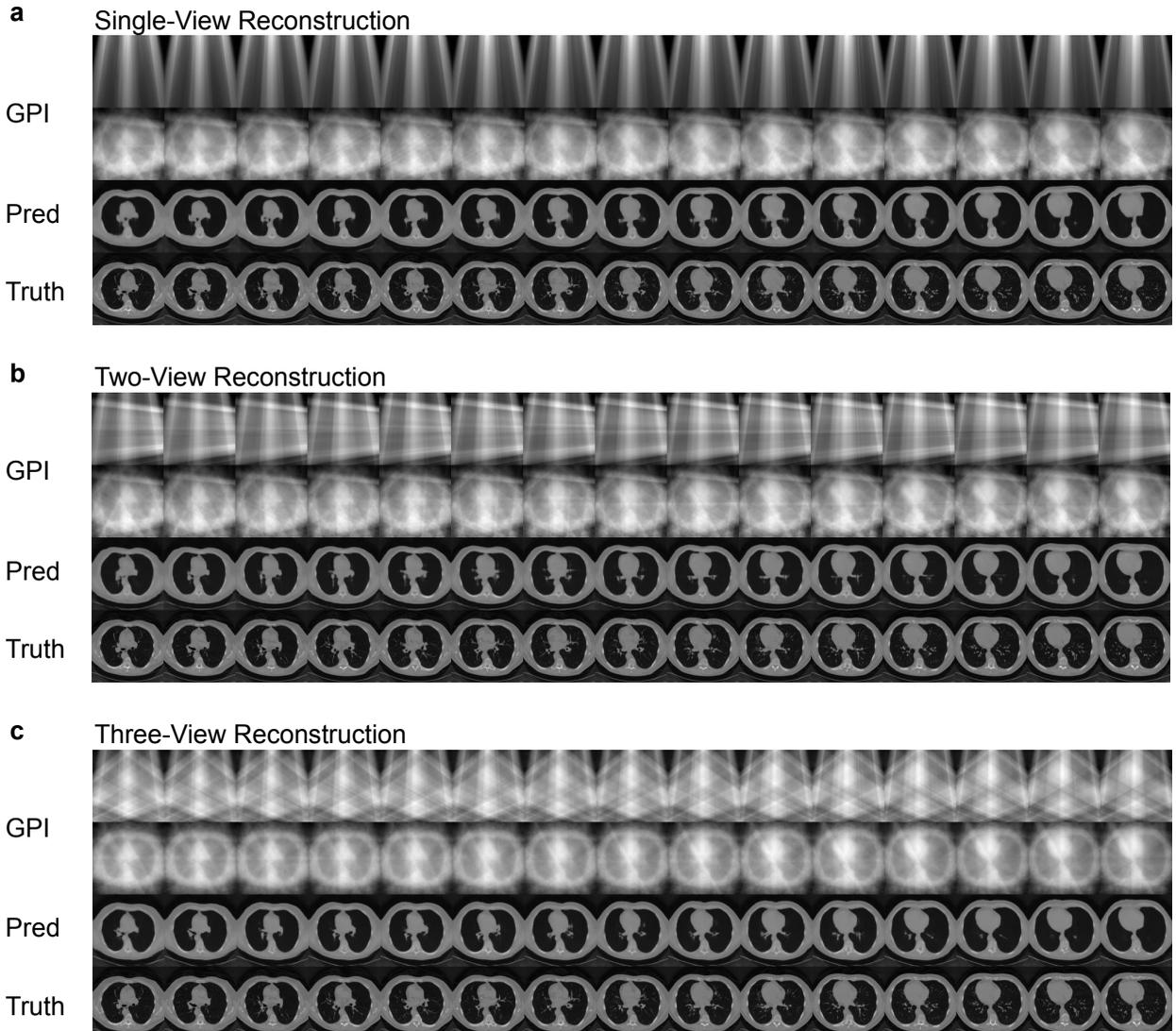

Fig. 8. Results of ultra-sparse reconstruction by GIIR model. Here, we show the cross-sectional slices of 3D CT images of the same testing sample from (a) single-view reconstruction, (b) two-view reconstruction, (c) three-view reconstruction, respectively. In each row, cross-sectional slices of ground-truth 3D images, predicted final 3D images, and GPIs, are shown when 1-3 projections are inputted to GIIR model. Two GPIs back-projected from ground-truth projections and generated projections separately serve as inputs to 3D image refinement network.

## 6 DISCUSSION

Imaging or visual representation of an object has two facets, image intensities and their spatial locations. The task of image reconstruction is to assign the right intensity to the right spatial point through the extraction of information buried in the sensor data. While deep learning has shown remarkable performance for image reconstruction, the existing data-driven methods are largely brute-force in nature as they rely totally on the inherent capability of deep learning to pin down the relationship between input and output data.

Many efforts have been devoted to improving the situation by, for example, development of physics-based learning to mitigate the limitations of data-driven approaches in image analysis and segmentation [59], but a systematic strategy in advancing the field, in particularly image reconstruction, remains elusive. The proposed GIIR provides a valuable framework for incorporating geometric priors of the underlying imaging system to enhance the robustness, data efficiency, and generalizability of deep learning-based image reconstruction.

In CT imaging, high-dimensional information of the internal anatomy is compressed and encoded into 2D projections. For artifacts-free image reconstruction, a sufficiently dense angular sampling of 2D projections that satisfies the Shannon-Nyquist criterion is required in traditional

**a** Baseline: w/o geometric prior

Testing Sample 1

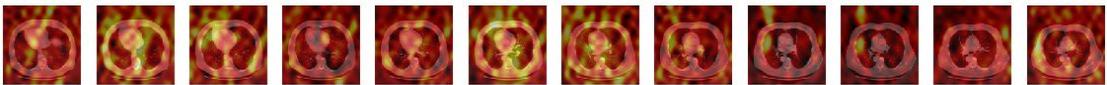

Testing Sample 2

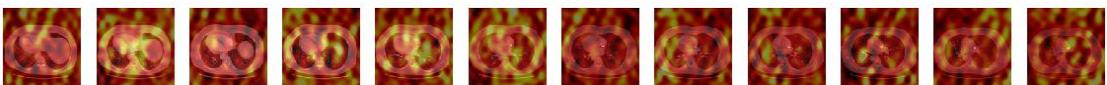

Testing Sample 3

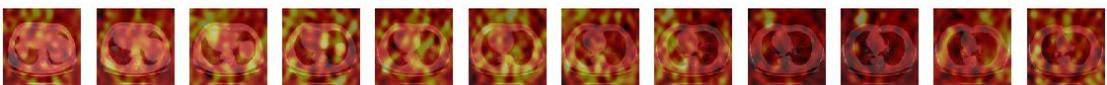

**b** GIIR: w/ geometric prior

Testing Sample 1

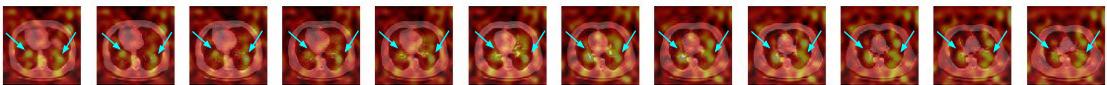

Testing Sample 2

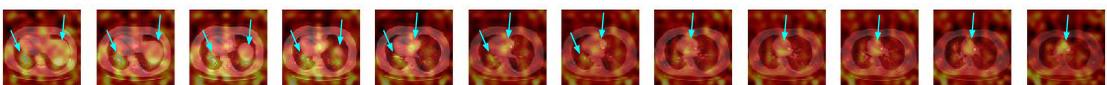

Testing Sample 3

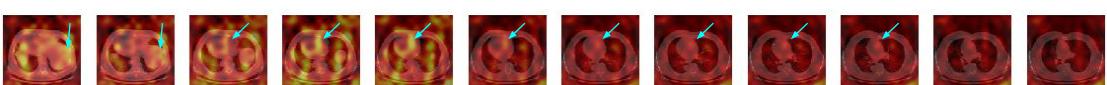

Fig. 9. Heatmaps of baseline and GIIR models for single-view reconstruction. (a) The feature maps for three testing samples companied with corresponding cross-sectional slices of CT images for the baseline model without geometric priors. (b) The heatmaps of the same three testing samples for GIIR model using geometric transformation to bridge the 2D-3D image domains.

model-based approaches. When the angular sampling is sparse, the information needed to recover the 3D image becomes inaccessible and other prior knowledge become necessary to solve this ill-posed inverse problem. Conventional sparse reconstruction, like compressed sensing [26], [60] [61], [62], [63], [64], introduces a regularization prior to encourage some presumed characteristics in the resultant image. Deep learning-based reconstruction [11], [12], [39], [65], [66], [67] gains knowledge by learning from large-scale training data.

The insight brought up here enables us to take advantage of unique properties of deep learning technique and geometry priors in image reconstruction. GIIR bridges the dimensionality gap between the 2D projection and 3D image domains via priors, which relieves the system from performing the non-trivial reasoning on the geometric relationship of the structures and allows the model to focus on learning semantic unknowns as visualized in Fig. 9. Our results demonstrate the power of incorporating prior knowledge from physical geometry, specifically for the application of ultra-sparse reconstruction. As a consequence of informing the deep neural networks with the geometry of the imaging system, the proposed GIIR demonstrates marked capability in reconstructing the correct 3D anatomic structure, even only when a single or a few projections are given. Note that the key insight of integrating geometry priors with deep learning in the proposed GIIR framework is a general model-agnostic approach, which is not limited by the specific network configuration. In other words, the specific network architecutres and loss objectives for the 2D projection synthesis network and the 3D image refinement network can be changed to different model choices.

The dual-domain learning in both 2D projection domain and 3D image domain is essential for sparse-view CT reconstruction. In most existing methods such as [35], the general insight is to use the conventional CT reconstruction approach such as FBP as a pre-processing step to obtain an initialization image and then refine the image by deep network. [66] [67] introduce dual-domain learning for solving metal artifacts problem in 2D CT reconstruction through sinogram interpolation but does not discuss how to conduct 2D projection domain learning for 3D CT reconstruction problem especially with ultra-sparse view. In this work, we propose a unified dual-domain learning framework for 3D CT reconstruction, synergistically combining both projection view synthesis and geometry-embedded image reconstruction. The underlying insight is that the dual-domain learning should be performed in both 2D projection domain and 3D image domain, where the geometry priors are introduced to bridge the dimensionality gap between 2D projection domain and 3D image domain through a deterministic back-projection transformation. In this way, the view synthesis module in 2D projection domain can help to relieve the burden in the image refinement. We observe that the introduction of projection synthesis, which helps to fill up the missing information in the 2D projection domain, is critical for single- or two-view reconstruction.

Ultra-sparse CT imaging strategy presented above had the potential to mitigate motion artifacts and make real-time volumetric CT imaging a reality, which promises to revolutionize many clinical image-guided interventional procedures such as radiation therapy and needle biopsy. Significant imaging dose reduction represents another potential benefit of the GIIR framework. The potential risk of cancer and genetic damage of X-ray CT has continued to attract public attention, and it is highly desirable to reduce the radiation dose in accordance with as low as reasonably achievable (ALARA) principle without compromising the image quality. The GIIR ultra-sparse CT imaging enables volumetric imaging with significantly reduced radiation dose.

In this study, we develop and evaluate the feasibility of GIIR using lung CT image reconstruction with ultra-sparse sampling data. The approach can be extended straightforwardly to other body sites, such as head and neck, abdomen, by using the training data acquired from the corresponding treatment sites. Of note, compared to these relatively stationary sites, lung CT image is much challenging due to the respiratory motion and it therefore poses a much urgent need for real-time volumetric lung CT imaging. However, there is still a long way to transfer this technique to real product for clinical diagnosis and the benefit of GIIR is also application specific. For example, for applications in radiation therapy, the reconstructed volumetric image could be used for patient positioning with much reduced dose. While for a subject with low-contrast lesion, sampling with a higher view number may be helpful to improve the lesion detectability. Beyond CT imaging, the general insight of integrating geometry prior with deep learning should be generalizable to a wide range of imaging problems such as other imaging modalities (e.g. PET, MRI).

Finally, the implications of the proposed strategy of prior geometric knowledge-informed deep learning goes much beyond image reconstruction. The general insights gained here should be applicable to many related disciplines involving geometric information, such as 3D scene reconstruction, view synthesis, video processing, robotic manipulation and drug design. Along this line of research, recent research works show promising results in various computer vision tasks [14], [18]. Coupled with our systematic study here, we believe that knowledge-integrated deep learning models have significant promise in advancing artificial intelligence to a new level.

## 7 CONCLUSION

A neural network is essentially an information processor and this process should be supplied with prior knowledge whenever available to maximally exploit the capacity of the data-driven technique. For a large cohort of problems, in which geometric transformation is part of the mapping function between the input and output, integration of the underlying geometric relationship would simplify the modeling process and enhance the robustness and generalizability of deep neural networks.

Specific to CT image reconstruction, we have shown that the process to unfold input sensor data into 3D space is made much more reliable with integration of known geometric information of the system. Technically, we have established a mechanism of integrating geometric information with deep learning-based reconstruction. The performance of the GIIR model for ultra-sparse sampling CT imaging has been demonstrated. The proposed strategy sheds important insights on the roles of prior knowledge in deep learning and illustrates how geometric prior knowledge helps the model to focus on learning other sematic unknowns to gain a better comprehension of the system.

Although this work is mainly focused on CT imaging, the general insight and concept of integrating geometric priors with deep learning could be extended to other imaging modalities, such as 3D object reconstruction in photography imaging, PET and MRI. Practically, this work may open new opportunities for numerous practical applications of data-driven image reconstruction.

## ACKNOWLEDGMENT

The authors acknowledge the National Cancer Institute and the Foundation for the National Institutes of Health, and their critical role in the creation of the free publicly available LIDC/IDRI Database used in this study. The authors acknowledge the funding supports from Stanford Bio-X Bowes Graduate Student Fellowship, National Cancer Institute (1R01CA227713), Google Faculty Research Award, and Human-Centered Artificial Intelligence (HAI) of Stanford University.

## CONFLICT OF INTEREST

A provisional patent application for the reported work has been filed. L.S and L.X. are co-inventors on the patent application based on this work. Licensing of the reported technique is managed by the Office of Technology Licensing (OTL) of Stanford University (Ref. Docket #S21-041).

## REFERENCES

[1] D. Heaven, "Why deep-learning AIs are so easy to fool," *Nature* 574, 163–166 (2019).

[2] M. Hutson, "Has artificial intelligence become alchemy?" *Science* 360, 478–478 (2018).

[3] S. G. Finlayson, J. D. Bowers, J. Ito, J. L. Zittrain, A. L. Beam, and I. S. Kohane, "Adversarial attacks on medical machine learning," *Science*, 363(6433):1287–1289 (2019).

[4] S. Ravishankar, J. C. Ye, and J. A. Fessler, "Image reconstruction: From sparsity to data-adaptive methods and machine learning," *Proceedings of the IEEE* 108(1), 86-109 (2019).

[5] D. Rueckert, and J. A. Schnabel, "Model-based and data-driven strategies in medical image computing," *Proceedings of the IEEE*, 108(1), 110-124 (2019).

[6] L. Xing, M. Giger and J. Min, "Artificial Intelligence in Medicine: Technical Basis and Clinical Applications," (Academic Press, Longdon, UK), (2020).

[7] A. Krizhevsky, I. Sutskever and G. E. Hinton, "Imagenet classification with deep convolutional neural networks," *Advances in Neural Information Processing Systems*, 1097-1105 (2012).

[8] K. He, X. Zhang, S. Ren and J. Sun, "Deep residual learning for image recognition," In *Proceedings of the IEEE Conference on Computer Vision and Pattern Recognition (CVPR)*, 770-778 (2016).

[9] L. Shen, et al. "Deep learning with attention to predict gestational age of the fetal brain," In *Machine Learning for Health Workshop of Advances in Neural Information Processing Systems* (2018).

[10] B. Zhu, J. Z. Liu, S. F. Cauley, B. R. Rosen and M. S. Rosen, "Image reconstruction by domain-transform manifold learning," *Nature* 555, 487-492 (2018).

[11] L. Shen, W. Zhao, and L. Xing, "Patient-specific reconstruction of volumetric computed tomography images from a single projection view via deep learning," *Nature Biomedical Engineering* 3(11), 880-888 (2019).

[12] X. Ying, et al. "X2CT-GAN: reconstructing CT from biplanar X-rays with generative adversarial networks," In *Proceedings of the IEEE Conference on Computer Vision and Pattern Recognition (CVPR)*, 10619-10628 (2019).

[13] L. H. Gilpin, D. Bau, B. Z. Yuan, A. Bajwa, M. Specter, and L. Kagal, "Explaining explanations: An overview of interpretability of machine learning," In *2018 IEEE 5th International Conference on Data Science and Advanced Analytics*, 80-89 (2018).

[14] H. Y. F. Tung, R. Cheng, and K. Fragkiadaki, "Learning spatial common sense with geometry-aware recurrent networks," In *Proceedings of the IEEE Conference on Computer Vision and Pattern Recognition (CVPR)*, 2595-2603 (2019).

[15] M. Tatarchenko, S. R. Richter, R. Ranftl, Z. Li, V. Koltun, and T. Brox, "What do single-view 3D reconstruction networks learn," In *Proceedings of the IEEE Conference on Computer Vision and Pattern Recognition (CVPR)*, 3405-3414 (2019).

[16] H. W. Engl, M. Hanke and A. Neubauer, "Regularization of inverse problems," (Springer Science & Business Media, 1996).

[17] J. Wang, T. Li, H. Lu and Z. Liang, "Penalized weighted least-squares approach to sinogram noise reduction and image reconstruction for low-dose X-ray computed tomography," *IEEE Trans. Med. Imaging* 25, 1272-1283 (2006).

[18] S. Wu, C. Rupprecht and A. Vedaldi, "Unsupervised learning of probably symmetric deformable 3D objects from images in the wild," In *Proceedings of the IEEE Conference on Computer Vision and Pattern Recognition (CVPR)*, 1-10 (2020).

[19] H. Fan, H. Su, and L. J. Guibas. "A point set generation network for 3d object reconstruction from a single image," In *Proceedings of the IEEE Conference on Computer Vision and Pattern Recognition*, pp. 605-613 (2017).

[20] C. B. Choy, D. Xu, J. Gwak, K. Chen, and S. Savarese, "3d-r2n2: A unified approach for single and multi-view 3d object reconstruction," In *European Conference on Computer Vision*, pp. 628-644 (2016).

[21] X. Yan, et al. "Perspective transformer nets: Learning single-view 3d object reconstruction without 3d supervision," *Advances in Neural Information Processing Systems*, pp. 1696-1704 (2016).

[22] A. Kar, C. Häne, and J. Malik, "Learning a multi-view stereo machine," In *Advances in Neural Information Processing Systems*, pp. 365-376 (2017).

[23] X. Zhang, Z. Zhang, C. Zhang, J. Tenenbaum, B. Freeman, and J. Wu, "Learning to reconstruct shapes from unseen classes," In *Advances in neural information processing systems*, pp. 2257-2268 (2018).

[24] P. Henzler, N. J. Mitra, and T. Ritschel, "Escaping plato's cave: 3D shape from adversarial rendering," *Proceedings of the IEEE International Conference on Computer Vision*, pp. 9984-9993, (2019).

[25] J.F. Barrett and N. Keat, "Artifacts in CT: recognition and avoidance," *Radiographics*, 24(6), pp.1679-1691 (2004).

[26] G. H. Chen, J. Tang and S. Leng, "Prior image constrained compressed sensing (PICCS): a method to accurately reconstruct dynamic CT images from highly undersampled projection data sets," *Med. Phys.* 35, 660-663 (2008).

[27] G. Wang and M. Jiang, "Ordered-subset simultaneous algebraic reconstruction techniques (OS-SART*)", Journal of X-ray Science and Technology*, 12(3), pp.169-177 (2004).

[28] K. Mueller and R. Yagel, "Rapid 3-D cone-beam reconstruction with the simultaneous algebraic reconstruction technique (SART) using 2-D texture mapping hardware," *IEEE Transactions on Medical imaging*, 19(12), pp.1227-1237 (2000).

[29] K. Mueller, R. Yagel and J.J. Wheller, "Anti-aliased three-dimensional cone-beam reconstruction of low-contrast objects with algebraic methods," *IEEE Transactions on Medical Imaging*, 18(6), pp.519-537 (1999).

[30] J. A. Fessler and W. L. Rogers, "Spatial resolution properties of penalizedlikelihood image reconstruction: space-invariant tomographs," *IEEE Trans. Image Process* 5, 1346–1358 (1996).

[31] S. Ji, Y. Xue and L. Carin, "Bayesian compressive sensing," *IEEE Trans. Signal Process*, 56, 2346–2356 (2008).

[32] H.M. Hudson and R.S. Larkin, "Accelerated image reconstruction using ordered subsets of projection data," *IEEE Transactions on Medical Imaging*, 13(4), pp.601-609 (1994).

[33] E.Y. Sidky, C.M. Kao and X. Pan, "Accurate image reconstruction from few-views and limited-angle data in divergent-beam CT," *Journal of X-ray Science and Technology*, 14(2), pp.119-139 (2006).

[34] H. Chen, et al. "LEARN: Learned experts' assessment-based reconstruction network for sparse-data CT," *IEEE transactions on medical imaging*, *37*(6), pp.1333-1347 (2018).

[35] Z. Zhang, X. Liang, X. Dong, Y. Xie and G. Cao. "A sparse-view CT reconstruction method based on combination of DenseNet and deconvolution," *IEEE transactions on medical imaging*, *37*(6), pp.1407-1417 (2018).

[36] Y. Han and J.C. Ye. "Framing U-Net via deep convolutional framelets: Application to sparse-view CT," *IEEE transactions on medical imaging*, *37*(6), pp.1418-1429 (2018).

[37] W.A. Lin, H. Liao, C. Peng, X. Sun, J. Zhang, J. Luo, R. Chellappa and S.K. Zhou. "Dudonet: Dual domain network for ct metal artifact reduction," In *Proceedings of the IEEE/CVF Conference on Computer Vision and Pattern Recognition*, pp. 10512-10521 (2019).

[38] L. Yu, Z. Zhang, X. Li and L. Xing. "Deep sinogram completion with image prior for metal artifact reduction in CT images," *IEEE Transactions on Medical Imaging*, *40*(1), pp.228-238 (2020).

[39] T. Würfl, M. Hoffmann, V. Christlein, K. Breininger, Y. Huang, M. Unberath and A.K. Maier. "Deep learning computed tomography: Learning projection-domain weights from image domain in limited angle problems," *IEEE transactions on medical imaging*, *37*(6), pp.1454-1463 (2018).

[40] M. M. Bronstein, J. Bruna, Y. LeCun, A. Szlam, and P. Vandergheynst, "Geometric deep learning: going beyond euclidean data," *IEEE Signal Processing Magazine*, 34(4), 18-42 (2017).

[41] S. A. Eslami, et al. "Neural scene representation and rendering". *Science*, *360*(6394), 1204-1210 (2018).

[42] V. Sitzmann, M. Zollhöfer, and G. Wetzstein. "Scene representation networks: Continuous 3d-structure-aware neural scene representations," In *Advances in Neural Information Processing Systems*, pp. 1121-1132 (2019).

[43] B. Mildenhall, et al. "Nerf: Representing scenes as neural radiance fields for view synthesis," In *Proceedings of the European Conference on Computer Vision (ECCV)* (2020).

[44] S. Lombardi, et al. "Neural volumes: learning dynamic renderable volumes from images," *ACM Trans. Graph.* 38(4), 65 (2019)

[45] O. Wiles, G. Gkioxari, R. Szeliski, and J. Johnson. "Synsin: End-to-end view synthesis from a single image," In *Proceedings of the IEEE Conference on Computer Vision and Pattern Recognition*, pp. 7467-7477 (2020).

[46] X. Huang, M.-Y. Liu, S. Belongie, and J. Kautz, "Multimodal unsupervised image-to-image translation," in *Proceedings of the European Conference on Computer Vision (ECCV)*, pp. 172–189 (2018).

[47] L. Shen, et al. "Multi-domain image completion for random missing data with representational disentanglement," In *IEEE Transactions on Medical Imaging* (2020).

[48] O. Ronneberger, P. Fischer, and T. Brox, "U-net: Convolutional networks for biomedical image segmentation," In *International Conference on Medical Image Computing and Computer-Assisted Intervention (MICCAI)*, 234-241 (2015).

[49] H. Seo, C. Huang, M. Bassenne, R. Xiao and L. Xing, "Modified U-Net (mU-Net) with incorporation of object-dependent high level features for improved liver and liver-tumor segmentation in CT images," *IEEE Trans. Med. Imaging* 39(5), 1316-1325 (2019).

[50] V. Nair and G. E. Hinton, "Rectified Linear Units Improve Restricted Boltzmann Machines," In *Proceedings of the 27th International Conference on Machine Learning (ICML)*, 807-814 (2010).

[51] Y. Wu, and K. He, "Group normalization," In *Proceedings of the European Conference on Computer Vision (ECCV)*, 3-19, (2018).

[52] SG. Armato III, et al. "The lung image database consortium (LIDC) and image database resource initiative (IDRI): a completed reference database of lung nodules on CT scans," *Medical physics* 38(2), 915-931 (2011).

[53] SG. Armato III, et al. "Data From LIDC-IDRI". *The Cancer Imaging Archive* (2015).

[54] K. Clark, et al. "The Cancer Imaging Archive (TCIA): Maintaining and Operating a Public Information Repository," *Journal of Digital Imaging*, 26(6), 1045-1057 (2013).

[55] R. Timmerman and L. Xing, "Image Guided and Adaptive Therapy," (Lippincott, Williams, and Wilkins, Baltimore, MD, 2009).

[56] A. Paszke, et al. "Automatic differentiation in pytorch," In *Proc. 30th Conference on Advances in Neural Information Processing Systems Autodiff Workshop* (2017).

[57] D. P. Kingma, and, J. Ba, "Adam: A method for stochastic optimization," In *Proceedings of the International Conference on Learning Representations (ICLR)* (2015).

[58] Z. Wang, A. C. Bovik, H. R. Sheikh and E. P. Simoncelli. "Image quality assessment: from error visibility to structural similarity," *IEEE Trans. Image Process.* 13, 600-612 (2004).

[59] J. R. Clough, et al. "Explicit topological priors for deep-learning based image segmentation using persistent homology," In *International Conference on Information Processing in Medical Imaging*, 16-28 (2019).

[60] E. J. Candes, J. K. Romberg and T. Tao, "Stable signal recovery from incomplete and inaccurate measurements," *Commun. Pure Appl. Math.* 59, 1207-1223 (2006).

[61] M. Lustig, D. Donoho and J. M. Pauly. "Sparse MRI: The application of compressed sensing for rapid MR imaging," *Magn. Reson. Med.* 58, 1182-1195 (2007).

[62] E. Y. Sidky and Pan, X, "Image reconstruction in circular cone-beam computed tomography by constrained, total-variation minimization," *Phys. Med. Biol.* 53, 4777-4807 (2008).

[63] H. Yu and G. Wang, "Compressed sensing based interior tomography," *Phys. Med. Biol.* 54, 2791-2805 (2009).

[64] K. Choi et al. "Compressed sensing based cone-beam computed tomography reconstruction with a first-order method," *Med. Phys*. 37, 5113-5125 (2010).

[65] K. H. Jin, M. T. McCann, E. Froustey, and M. Unser, "Deep convolutional neural network for inverse problems in imaging," *IEEE Trans. Image Process*. 26(9), 4509-4522 (2017).

[66] J. Adler, and O. Öktem, "Learned primal-dual reconstruction," *IEEE Trans. Med. Imaging* 37(6), 1322-1332 (2018).

[67] A. K. Maier, et al. Learning with known operators reduces maximum error bounds. *Nature Machine Intelligence* 1(8), 373-380 (2019).